\documentclass[runningheads]{llncs}

\usepackage[T1]{fontenc}

\usepackage{graphicx}
\usepackage{xcolor}
\usepackage{hyperref}

\usepackage{caption}
\usepackage{subcaption} 

\begin{document}

\title{CitiLink: Enhancing Municipal Transparency and Citizen Engagement through Searchable Meeting Minutes} 
\titlerunning{CitiLink: Enhancing Access to Municipal Meeting Minutes}
%
\author{
Rodrigo~Silva\inst{1,3}\orcidID{0009-0009-3426-186X} \and
José~Evans\inst{2,3}\orcidID{0009-0003-6408-5103} \and
José~Isidro\inst{2,3}\orcidID{0009-0000-6071-9138} \and
Miguel~Marques\inst{1,3}\orcidID{0009-0002-7934-0173} \and
Afonso~Fonseca\inst{1,3}\orcidID{0009-0006-2160-4315} \and
Ricardo~Morais\inst{2}\orcidID{0000-0001-8827-0299} \and
João~Canavilhas\inst{1}\orcidID{0000-0002-2394-5264} \and
Arian~Pasquali\inst{3}\orcidID{0000-0002-3487-9397} \and
Purificação~Silvano\inst{2,3}\orcidID{0000-0001-8057-5338} \and
Alípio~Jorge\inst{2,3}\orcidID{0000-0002-5475-1382} \and
Nuno~Guimarães\inst{2,3}\orcidID{0000-0003-2854-2891} \and
Sérgio~Nunes\inst{2,3}\orcidID{0000-0002-2693-988X} \and
Ricardo~Campos\inst{1,3}\orcidID{0000-0002-8767-8126}
}
\authorrunning{R.~Silva et al.}

\institute{
University of Beira Interior, Covilhã, Portugal \\
\email{\{rd.silva, ricardo.campos\}@ubi.pt}
\and
University of Porto, Porto, Portugal
\and
INESC TEC, Porto, Portugal \\
}
\maketitle              
\begin{abstract}
City council minutes are typically lengthy and formal documents with a bureaucratic writing style. Although publicly available, their structure often makes it difficult for citizens or journalists to efficiently find information. In this demo, we present CitiLink, a platform designed to transform unstructured municipal meeting minutes into structured and searchable data, demonstrating how NLP and IR can enhance the accessibility and transparency of local government. The system employs LLMs to extract metadata, discussed subjects, and voting outcomes, which are then indexed in a database to support full-text search with BM25 ranking and faceted filtering through a user-friendly interface. The developed system was built over a collection of 120 minutes made available by six Portuguese municipalities. To assess its usability, CitiLink was tested through guided sessions with municipal personnel, providing insights into how real users interact with the system. In addition, we evaluated Gemini’s performance in extracting relevant information from the minutes, highlighting its effectiveness in data extraction.


\end{abstract}
\setcounter{footnote}{0}
\section{Introduction}
Municipal councils are often subject to transparency metrics designed to assess how effectively local authorities disclose information and keep citizens informed~\cite{araujo2016local}. A common practice supporting these metrics is the release of meeting minutes. While these records contribute to transparency~\cite{daCruz02072016}, their formal, bureaucratic style and administrative jargon~\cite{orebe2021linguistic} make them difficult to understand, posing barriers for citizens or journalists, for whom reading minutes becomes a particularly time-consuming task~\cite{LocalView2023}. Transforming unstructured minutes into structured data presents therefore an opportunity to facilitate efficient information access and organized navigation of council records. Although prior research has explored video recordings of council meetings~\cite{jiang2023natural,maxfield2022councils,vanwijk2025spoken}, written minutes remain largely underexplored, particularly in European Portuguese~\cite{rodrigues2010knowledge}. In this paper, we present CitiLink, a platform designed to address this gap by providing structured access to city council minutes. To populate the platform, we developed a pipeline that converts unstructured minutes into structured data, using LLMs with prompt engineering techniques. Our demo relies on 120 meeting minutes, manually anonymized with respect to personal data, and made available by 6 Portuguese municipalities: Alandroal, Campo Maior, Covilhã, Fundão, Guimarães, and Porto. The corpus spans diverse contexts and municipality sizes, with considerable variation in document length, formality, and subjects of discussion. For demonstration purposes, an English version of the dataset was created using DeepL translations, illustrating the system's multilingual capabilities. 

\section{System Overview}
\label{sec:sistema}
\begin{figure}[t]
    \centering
    \includegraphics[width=\linewidth]{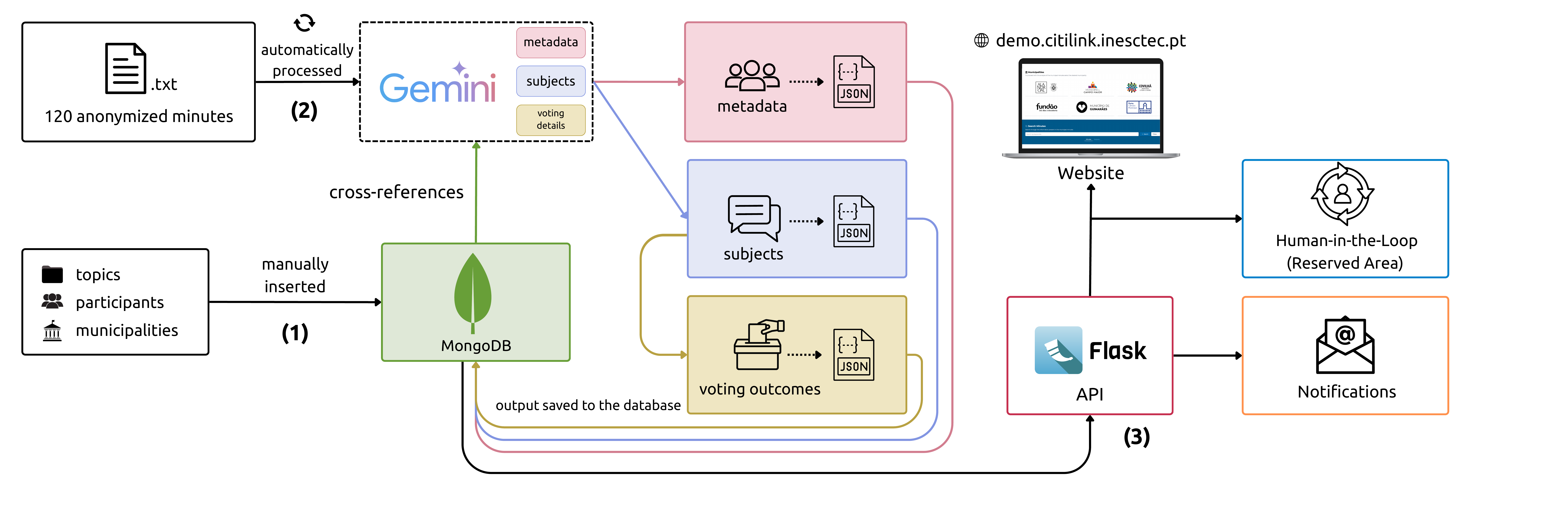}
    \caption{Architecture of the CitiLink system.}
    \label{fig:arq}
\end{figure}

The system consists of a pipeline that leverages an LLM (Gemini 2.0 Flash) to extract information from meeting minutes, a front-end web application with integrated IR features, and a restricted back-office where municipalities can upload minutes and validate extracted data, ensuring human-in-the-loop oversight. Users can also subscribe to a newsletter for updates. An overview of the system architecture is shown in Figure~\ref{fig:arq}. Each minute is provided to the LLM as plain text, which, using prompt engineering, extracts three layers of information: (1)~metadata (participants, location, date, type of meeting), (2)~subjects of discussion (e.g., ``Modification of traffic regulations on Avenida Central''), and (3)~voting outcomes (in favor, against, abstention). Extracted metadata are cross-referenced with predefined database collections (participants, municipalities, and topics) to ensure consistency. From the 120 meeting minutes, the system extracted 115 unique metadata elements (86 participants, 16 locations, 111 dates, 2 minute types), 3,079 subjects of discussion, and 24,040 votes (22,702 in favor, 161 against, 1,177 abstentions). All data are stored in MongoDB Atlas to support full-text and faceted search. The front-end, built with React, allows searching within minutes and subjects, or browsing records by municipality, while a Flask API provides access to the processed data. The prompts, processing pipeline code, evaluation guide, and Gemini-generated data are publicly available on GitHub\footnote{\url{https://github.com/inesctec/citilink-demo}}.

\section{Demonstration}
\label{sec:demo}
\begin{figure}[t]
    \centering
    \begin{subfigure}{0.25\textwidth}
        \centering
        \includegraphics[width=\linewidth]{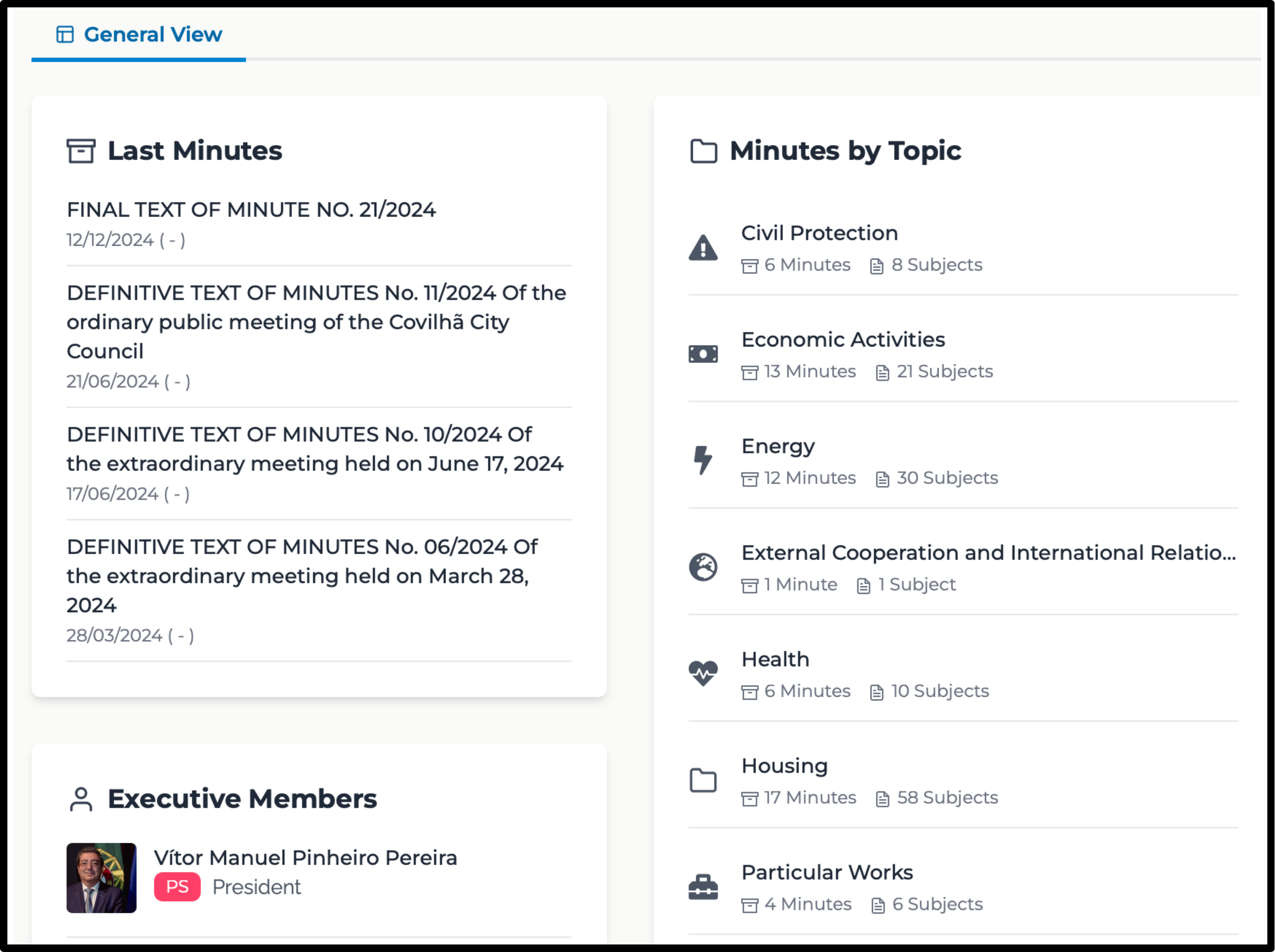}
        \caption{}
        \label{fig:general_view}
    \end{subfigure}%
    \hfill
       \begin{subfigure}{0.33\textwidth}
        \centering
        \includegraphics[width=\linewidth]{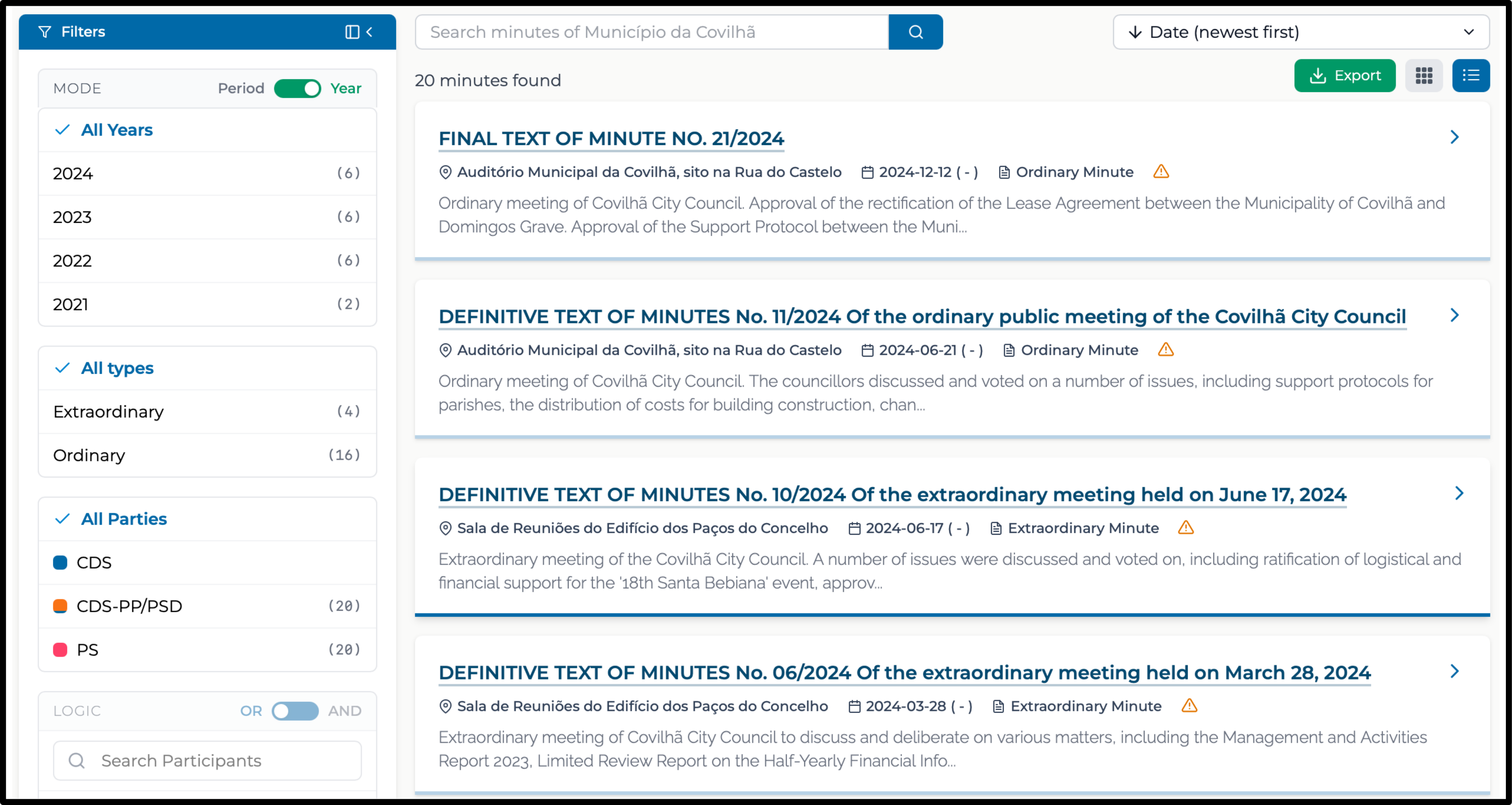}
        \caption{\small}
        \label{fig:list_view}
    \end{subfigure}%
    \hfill
    \begin{subfigure}{0.37\textwidth}
        \centering
        \includegraphics[width=\linewidth]{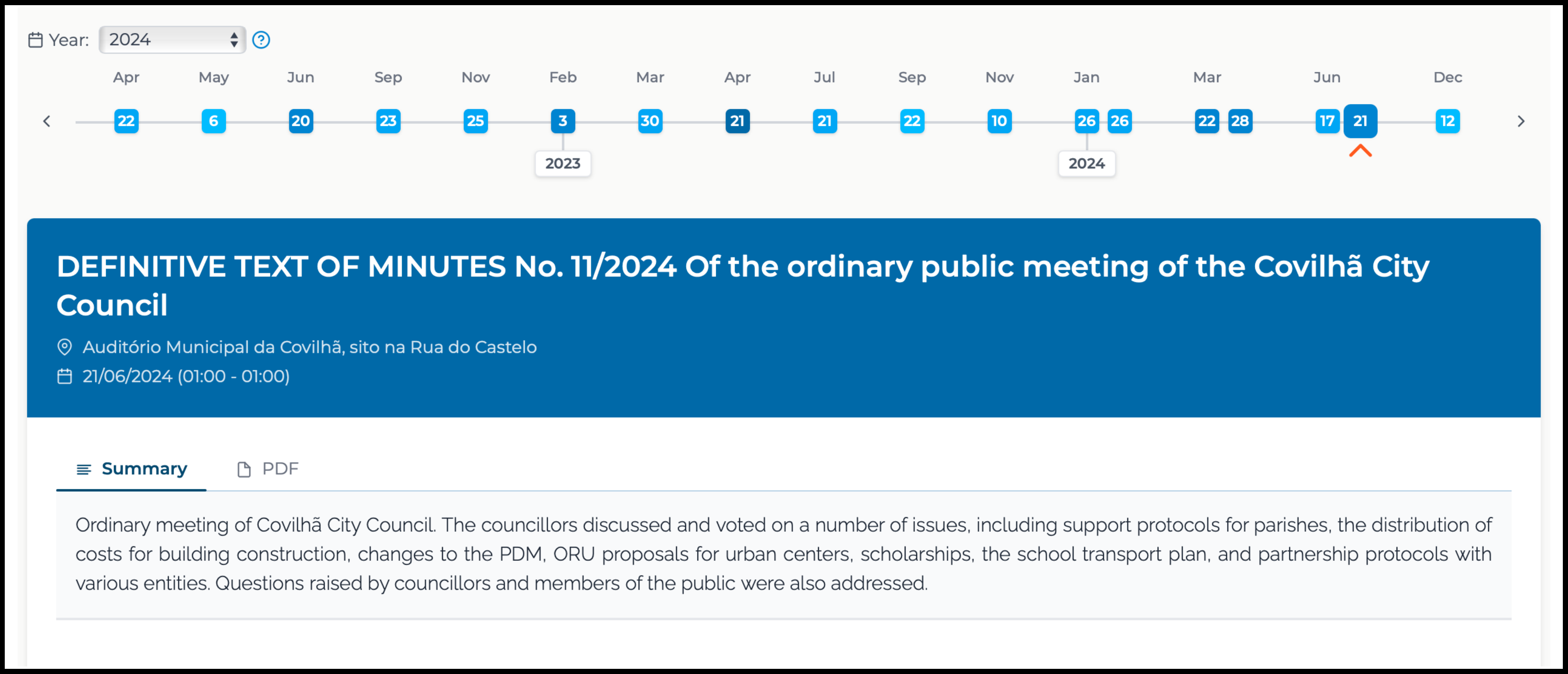}
        \caption{\small}
        \label{fig:timeline}
    \end{subfigure}
    \hfill
       \begin{subfigure}{0.5\textwidth}
        \centering
        \includegraphics[width=\linewidth]{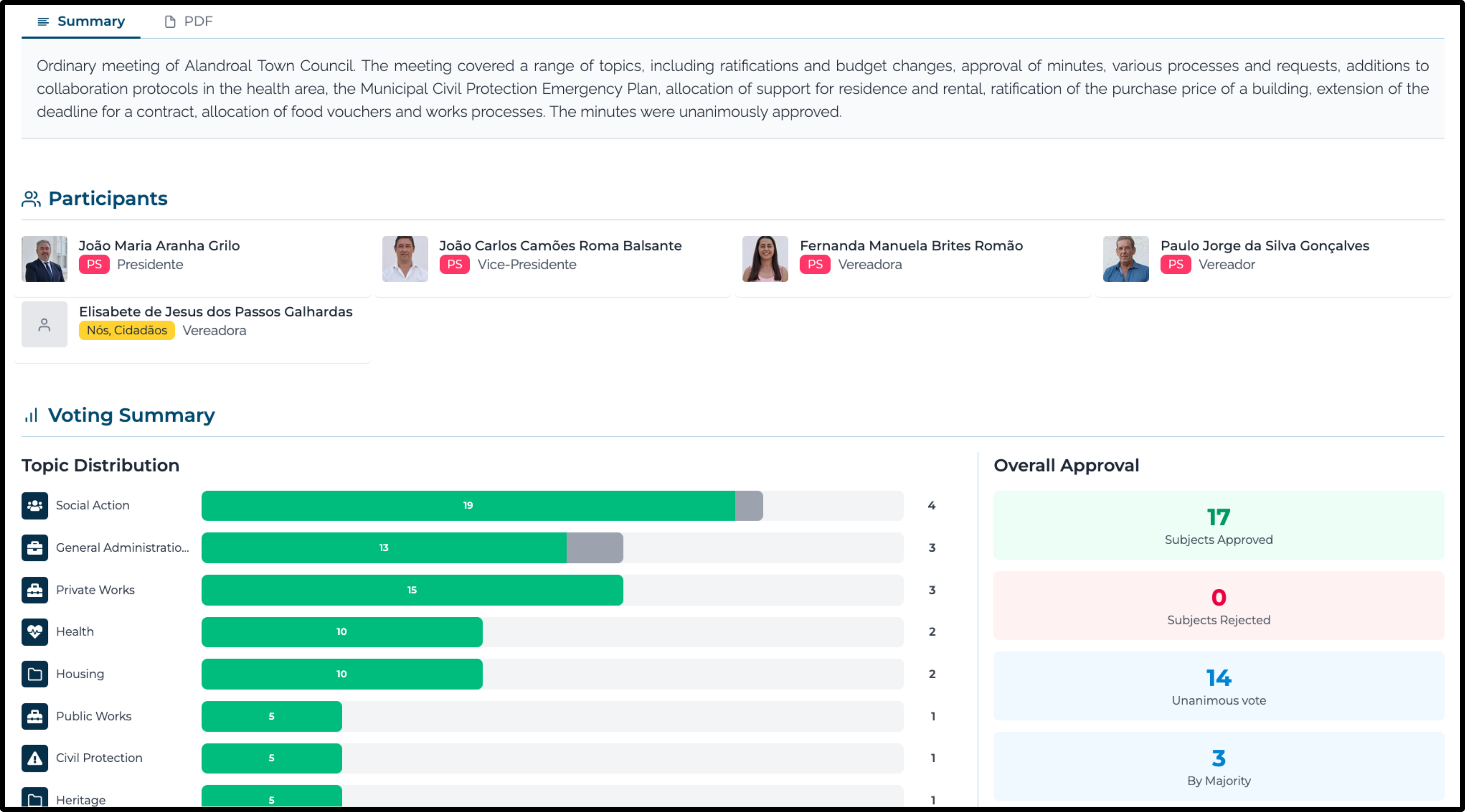}
        \caption{\small}
        \label{fig:pagina_ata}
    \end{subfigure}%
    \hfill
    \begin{subfigure}{0.45\textwidth}
    \centering
    \begin{minipage}{\linewidth}
        \centering\includegraphics[width=0.8\linewidth]{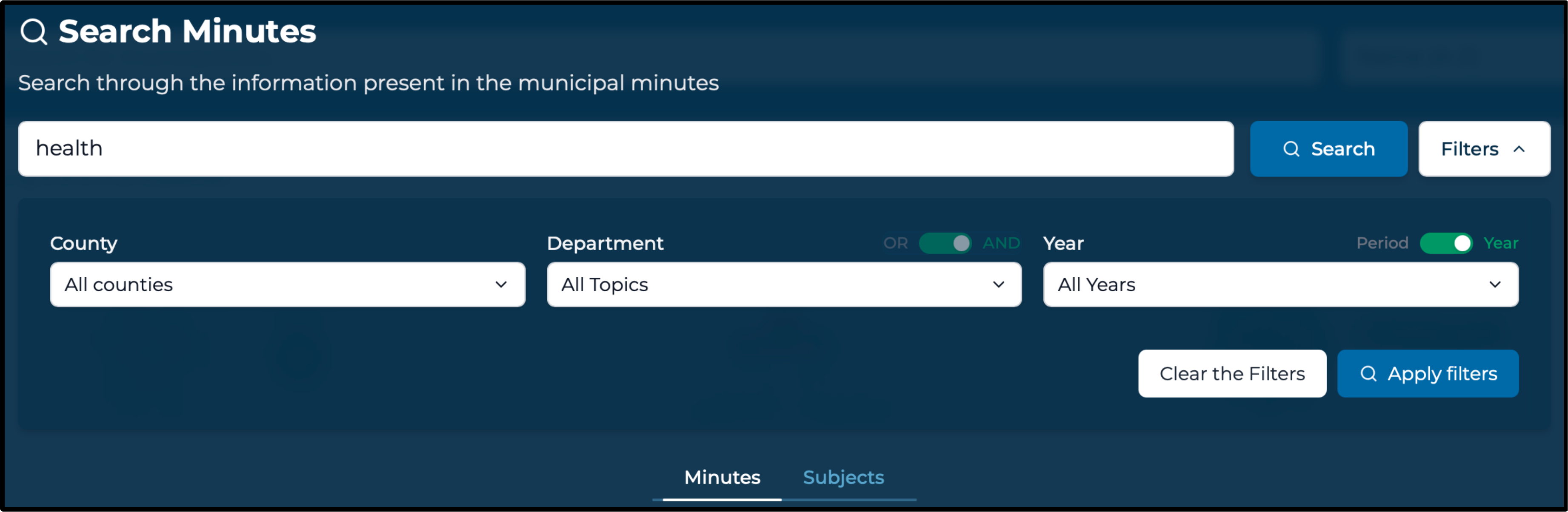}
        \caption{\small}
        \label{fig:search}
    \end{minipage}
    \vspace{0.5em} 
    \begin{minipage}{\linewidth}
        \centering\includegraphics[width=0.8\linewidth]{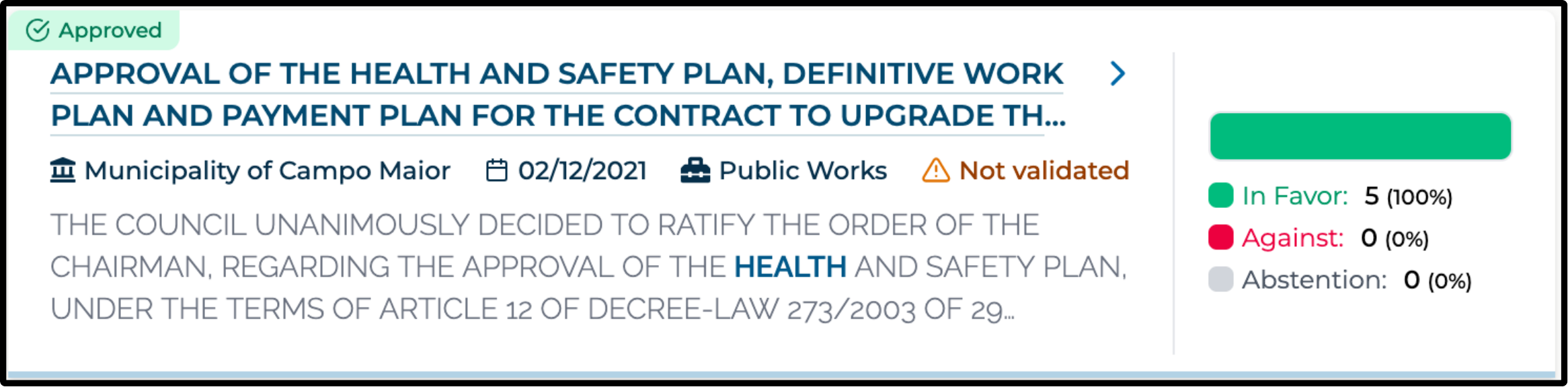}
        \caption{\small}
        \label{fig:assunto}
    \end{minipage}
\end{subfigure}
    
    \caption{Main components of the CitiLink web interface: (a) Municipality Overview; (b) List-based view; (c) Timeline view; (d) Minute page; (e) Search interface; (f) Search results interface.}
    \label{fig:overall}
\end{figure}

The CitiLink platform\footnote{\url{https://demo.citilink.inesctec.pt/en}, password: \texttt{ecir2026}} provides a web interface for exploring and interacting with the retrieved data. On the main page, users see cards for the six municipalities alongside a search section. 
Selecting a municipality presents a general overview of the most recent meeting minutes, executive members, and minutes grouped by discussion topics (as shown in Figure~\ref{fig:general_view}). In addition to this overview, users can access a list-based view for easy access to the minutes and faceted search by topic, party, or participant (as illustrated by Figure~\ref{fig:list_view}) or a timeline view for chronological exploration (as seen in Figure~\ref{fig:timeline}). By selecting a meeting minute, users can subsequently access a structured overview of its contents (as exemplified by Figure~\ref{fig:pagina_ata}), including metadata (e.g., participants), a voting summary, and detailed information on the subjects of discussion (omitted from the figure due to space constraints), helping users understand council decisions without needing to read the full minutes. The platform also supports full-text search (as shown in Figure~\ref{fig:search}), allowing both exploratory navigation and targeted retrieval. For example, querying ``\textit{health}'' retrieves the most relevant associated subjects (as shown in Figure~\ref{fig:assunto}). A demonstration video of the platform is available on the demo website and GitHub, showcasing all its features. 

\section{Evaluation and Future Work}
\label{sec:eval}
To evaluate CitiLink’s usability and effectiveness, preliminary assessments were conducted with municipal personnel between July 23 and September 4, 2025. Eight testing sessions, each lasting approximately 50 minutes, involved administrative staff and individuals responsible for producing or reviewing meeting minutes, ensuring a diversity of professional backgrounds and technical experience. The evaluation followed a think-aloud methodology~\cite{10.5555/2821575,10.1007/978-3-031-83845-3_7}, in which participants verbalized their thoughts while interacting with the platform. During the sessions, participants performed ten guided tasks designed to assess navigation, functionality, and the identification of votes and discussion topics (the guide is available on the Github repository). The analysis of the sessions indicated that CitiLink has substantial potential to modernize access to municipal meeting minutes. All participants found navigation intuitive, with six noting they would use the platform again. Based on participants’ remarks, the demo was refined and improved accordingly, exemplifying a user-centered refinement of the system.
In addition to usability, we assessed the performance of Gemini 2.0 Flash on each of the three extraction layers (metadata, subjects of discussion, and voting outcomes) and their individual items. The evaluation relied on the 120 meeting minutes of the Portuguese version of the dataset\footnote{\url{https://github.com/inesctec/citilink-dataset}}. A description of the dataset is out of the scope of this demonstration paper; briefly, the minutes are annotated with multiple layers, including personal information, metadata, subjects, and voting outcomes, following a controlled annotation protocol. Each document was independently annotated by two trained annotators and subsequently validated by a curator, with inter-annotator agreement reported in prior work describing the dataset. The current corpus size reflects the benchmark used for validation, while the system architecture is designed to scale incrementally to larger municipal archives. The metadata achieved an overall macro F1 of 0.84 across the 6 municipalities, considering all defined metadata fields, indicating consistent performance across categories. For subjects, direct comparison between LLM-outputs and ground-truth annotations is challenging, as the LLM does not produce text offsets. To enable evaluation, each ground-truth subject was matched one-by-one to its most similar LLM-generated candidate using BERTimbau~\cite{souza2020bertimbau} embeddings and cosine similarity. ROUGE-L (0.31) and BLEU (0.21) indicate that the LLM captures aspects of content and structure, while wording differences reflect task variability. Finally, in the voting task, we evaluated Gemini's ability to identify councilor's voting position-favor, against, or abstained, by comparing its outputs with the ground truth (e.g., ``Councilman John voted in favor''). The model achieved a macro F1 of 0.67, reflecting the greater complexity of this task compared to structured metadata extraction. These results suggest room for improvement and motivate tests with alternative LLMs.
In future work, we plan to develop and integrate open-source language models across the different layers and expand evaluation through citizen and journalist feedback via questionnaires and focus groups, enabling a broader assessment of CitiLink’s usability and impact across diverse contexts.

\begin{credits}

\subsubsection*{Preprint and Version of Record}
This preprint has not undergone peer review (when applicable) or any post-submission improvements or corrections. The Version of Record of this contribution is published in {Advances in Information Retrieval. ECIR 2026. Lecture Notes in Computer Science, vol 16486. Springer, Cham.}, and is available online at https://doi.org/10.1007/978-3-032-21321-1\_28.

\subsubsection*{\ackname}
This work was funded within the scope of the project CitiLink, with reference 2024.07509.IACDC, which is co-funded by Component 5 - Capitalization and Business Innovation, integrated in the Resilience Dimension of the Recovery and Resilience Plan within the scope of the Recovery and Resilience Mechanism (MRR) of the European Union (EU), framed in the Next Generation EU, for the period 2021 - 2026, measure RE-C05-i08.M04 - ``To support the launch of a programme of R\&D projects geared towards the development and implementation of advanced cybersecurity, artificial intelligence and data science systems in public administration, as well as a scientific training programme,'' as part of the funding contract signed between the Recovery Portugal Mission Structure (EMRP) and the FCT - Fundação para a Ciência e a Tecnologia, I.P.~(Portuguese Foundation for Science and Technology), as intermediary beneficiary.\footnote{\url{https://doi.org/10.54499/2024.07509.IACDC}}
This work is also funded by national funds through FCT – Fundação para a Ciência e a Tecnologia, I.P., under the support UID/50014/2025.\footnote{\url{https://doi.org/10.54499/UID/50014/2025}}


\end{credits}

%
\bibliographystyle{splncs04}
\bibliography{ref}

@inproceedings{vanwijk2025spoken,
  author    = {Pepijn van Wijk and Maarten Marx},
  title     = {{Spoken Question Answering on Municipal Council Meetings}},
  booktitle = {Proceedings of the European Conference on Information Retrieval (ECIR 2025)},
  year      = {2025},
  organization = {IRLab, Informatics Institute, University of Amsterdam},
}

@article{maxfield2022councils,
  title={{Councils in action: Automating the curation of municipal governance data for research}},
  author={Maxfield Brown, Eva and Weber, Nicholas},
  journal={Proceedings of the Association for Information Science and Technology},
  volume={59},
  number={1},
  pages={23--31},
  year={2023},
  publisher={Wiley Online Library}
}

@article{LocalView2023,
  title={{LocalView, a database of public meetings for the study of local politics and policy-making in the United States}},
  author={Soubhik Barari and Tyler Simko},
  journal={Scientific Data},
  volume={10},
  number={1},
  pages={135},
  year={2023},
  publisher={Nature},
  issn={2052-4463}
}

@article{daCruz02072016,
author = {Nuno Ferreira da Cruz and António F. Tavares and Rui Cunha Marques and Susana Jorge and Luís de Sousa},
title = {{Measuring Local Government Transparency}},
journal = {Public Management Review},
volume = {18},
number = {6},
pages = {866--893},
year = {2016},
publisher = {Routledge},
doi = {10.1080/14719037.2015.1051572},


URL = { 
    
        https://doi.org/10.1080/14719037.2015.1051572
    
    

},
eprint = { 
    
        https://doi.org/10.1080/14719037.2015.1051572
    
    

}

}

@article{orebe2021linguistic,
  title={A linguistic-stylistic analysis of selected aspects of minutes of meeting},
  author={Orebe, Oluwabukola O},
  journal={Journal of Language Teaching and Research},
  volume={12},
  number={2},
  pages={286--292},
  year={2021},
  publisher={Academy Publication Co., Ltd.}
}

@article{jiang2023natural,
  title={Natural language processing adoption in governments and future research directions: A systematic review},
  author={Jiang, Yunqing and Pang, Patrick Cheong-Iao and Wong, Dennis and Kan, Ho Yin},
  journal={Applied Sciences},
  volume={13},
  number={22},
  pages={12346},
  year={2023},
  publisher={MDPI}
}

@article{rodrigues2010knowledge,
  title={Knowledge extraction from minutes of portuguese municipalities meetings},
  author={Rodrigues, M{\'a}rio and Dias, Gon{\c{c}}alo Paiva and Teixeira, Ant{\'o}nio},
  journal={Proc. of the FALA},
  pages={58},
  year={2010}
}

@article{araujo2016local,
  title={Local government transparency index: determinants of municipalities’ rankings},
  author={Araujo, Joaquim Filipe Ferraz Esteves de and Tejedo-Romero, Francisca},
  journal={International Journal of Public Sector Management},
  volume={29},
  number={4},
  pages={327--347},
  year={2016},
  publisher={Emerald Group Publishing Limited}
}

@book{10.5555/2821575,
author = {Nielsen, Jakob},
title = {Usability Engineering},
year = {1994},
isbn = {9780080520292},
publisher = {Morgan Kaufmann Publishers Inc.},
address = {San Francisco, CA, USA}
}

@InProceedings{10.1007/978-3-031-83845-3_7,
author="Zhao, Yihang
and Pe{\~{n}}uela, Albert Mero{\~{n}}o
and Simperl, Elena",
editor="Pl{\'a}cido da Silva, Hugo
and Cipresso, Pietro",
title="User Experience in Dataset Search",
booktitle="Computer-Human Interaction Research and Applications",
year="2025",
publisher="Springer Nature Switzerland",
address="Cham",
pages="113--130",
isbn="978-3-031-83845-3"
}

@inproceedings{souza2020bertimbau,
  author    = {F{\'a}bio Souza and
               Rodrigo Nogueira and
               Roberto Lotufo},
  title     = {{BERT}imbau: pretrained {BERT} models for {B}razilian {P}ortuguese},
  booktitle = {9th Brazilian Conference on Intelligent Systems, {BRACIS}, Rio Grande do Sul, Brazil, October 20-23 (to appear)},
  year      = {2020}
}

\end{document}